\documentclass[conference]{IEEEtran}
\usepackage{blindtext, graphicx}
\usepackage{subfigure}
\usepackage{amsmath}
\usepackage{amssymb}
\usepackage[square,sort,compress,comma,numbers]{natbib}
\usepackage{amsfonts}
\usepackage{verbatim}

\begin{document}

\title{Design of a Time Delay Reservoir Using Stochastic Logic:  A Feasibility Study}

\author{\IEEEauthorblockN{Cory Merkel}
\IEEEauthorblockA{Information Directorate\\
Air Force Research Laboratory\\
Rome, NY 13441\\
Email: cory.merkel.1@us.af.mil}}

\maketitle

\begin{abstract}
This paper presents a stochastic logic time delay reservoir design.  The reservoir is analyzed using a number of metrics, such as kernel quality, generalization rank, performance on simple benchmarks, and is also compared to a deterministic design.  A novel re-seeding method is introduced to reduce the adverse effects of stochastic noise, which may also be implemented in other stochastic logic reservoir computing designs, such as echo state networks.  Benchmark results indicate that the proposed design performs well on noise-tolerant classification problems, but more work needs to be done to improve the stochastic logic time delay reservoir's robustness for regression problems.
\end{abstract}


\section{Introduction}

Reservoir computing (RC) is proving to be a powerful machine learning technique for regression, classification, and forecasting of time series data.  Introduced in the early 2000s by Jaeger \cite{jaeger2001echo} and Maass \cite{maass2002real}, RC is a type of neural network with an untrained recurrent hidden layer called a \textit{reservoir}.  A major computational advantage of RC is that the output of the network can be trained on the reservoir states using simple regression techniques, without the need for backpropagation.  In the last decade and a half, RC has been successful in a number of wide-ranging applications domains such as image classification \cite{Woodward2011}, biosignal processing \cite{Kudithipudi2015}, and optimal control \cite{Tsai2010}.  In some domains, RC has outperformed state-of-the-art techniques and is often easier to implement than methods such as Kalman filtering or long short term memory.  Beyond its computational advantages, one of the main attractions of RC is that it can be implemented efficiently in hardware with low area and power overheads. 

Today, there are three major categories of RC.  The first is echo state networks (ESNs) \cite{jaeger2001echo}, where reservoirs are implemented using a recurrent network of continuous (e.g. logistic sigmoid) neurons.  The second category, referred to as liquid state machines (LSMs) \cite{maass2002real} utilizes recurrent connections of spiking (e.g. leaky integrate and fire) neurons.  A challenge in both of these categories is routing.  A reservoir with $H$ neurons will have up to $H^{2}$ connections, potentially creating a large area and power overhead.  A third category of RC called time delay reservoirs (TDR) \cite{Appeltant2011} avoids this overhead by time multiplexing resources.  TDRs utilize a single neuron and a delayed feedback to create reservoirs with either a chain topology or even full connectivity (see Supplemental Material of \cite{Appeltant2011}).

Besides a reduction in routing overhead, TDRs have two key advantages over ESNs and LSMs.  First, adding additional neurons to the reservoir is trivial and amounts to increasing the delay in the feedback loop.  Second, TDRs can use any dynamical system to implement their activation function and can easily be modeled via delay differential equations.  This second point is particularly useful since it means that TDRs can be implemented using a variety of technologies.  For example, in \cite{Appeltant2011}, Appeltant et al. used a Mackey-Glass oscillator, which models a number of physiological processes (e.g. blood circulation), as the non-linear node in a TDR.  In \cite{vinckier2015high}, a TDR is demonstrated using coherently driven passive optical cavity.  A TDR has also been implemented using a single XOR gate with delayed feedback \cite{Haynes2015}.  A common thread among all of these implementations is that they are analog and some, such as the photonic implementation, are still large prototypes that have yet to be integrated into a chip.  Aside from the higher cost and design effort for analog implementations, they are much more susceptible to noise, especially in RC, where the system operates on the edge of chaos.

Digital RC designs, and digital circuits in general, have much better noise immunity compared to analog implementations.  There have been a number of digital designs proposed for ESNs and LSMs, such as \cite{jin2016sso}, but digital TDR designs are presently scarce.  One example is given in \cite{Alomar2015}, where the authors have implemented a Mackey-Glass-type TDR on an FPGA.  One of the challenges with digital implementations is that the area cost can be high due to the requirement of multipliers for input weighting and implementation of the activation function.  This is especially true if high precision is required.  However, not all applications require high precision.  An alternative design approach to conventional digital logic is stochastic logic, where values are represented as stochastic bit streams and characterized by probabilities.  Stochastic logic has previously been used to implement ESNs \cite{Alomar2016, Verstraeten}.  In this work, we explore the feasibility of implementing TDRs with stochastic logic.  To the best of the author's knowledge, this is the first paper discussing TDR implementation with stochastic logic, and hopefully it will serve as a foundation for future research in this area.

The rest of this paper proceeds as follows:  Section \ref{section:background} provides background information on TDRs and stochastic logic.  Section \ref{section:design} presents the stochastic logic TDR designed in this work and discusses tuning of design parameters.  Section \ref{section:results} discusses the performance of the proposed TDR design on two benchmark tasks:  NARMA10 (regression) and sine/square wave discrimination (classification).  Section \ref{section:conclusions} concludes this work.

\section{Background}
\label{section:background}
\subsection{Time Delay Reservoirs}

\begin{figure}
\centering
\includegraphics[width=\columnwidth]{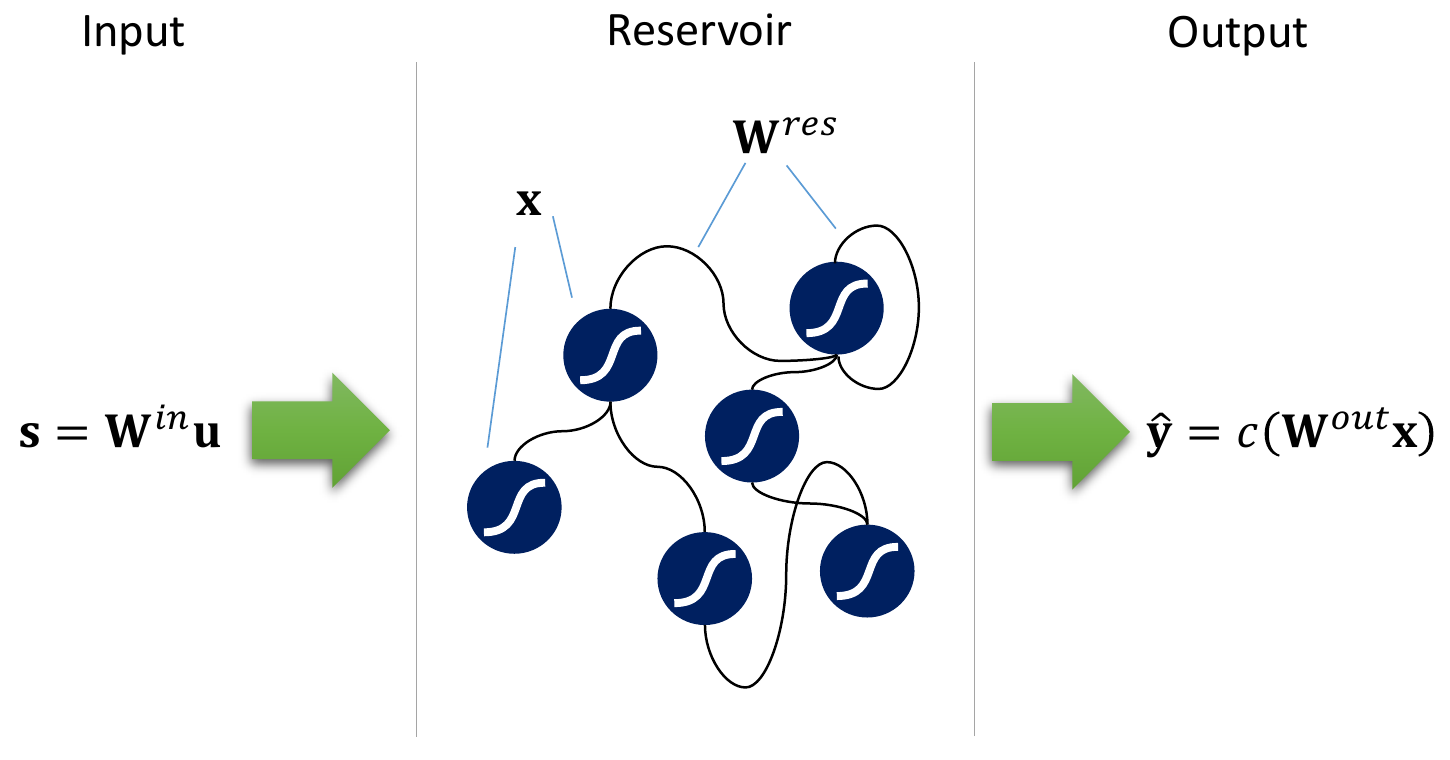}
\vspace{-5mm}
\caption{Overview of RC.  (a) Basic structure of an RC design, showing the three layers:  input, reservoir, and output.}
\label{fig:rc}
\end{figure}

RC makes use of a random recurrent neural network in order to regress, forecast, or classify time series data.  The basic structure of an RC is shown in Figure \ref{fig:rc}.  Time series inputs in the \textit{input layer} are multiplied by a random weight matrix $\mathbf{s}^{(p)}=\mathbf{W}^{in}\mathbf{u}^{(p)}$, and then used as inputs to the \textit{reservoir layer}.  Here, the index $p$ is used to denote a discrete timestep. Within the reservoir layer, there are a number of neurons (circles) that are connected with each other through a random weight matrix $\mathbf{x}^{(p+1)}=f\left(\mathbf{W}^{res}\mathbf{x}^{(p)}+\mathbf{s}^{(p)}\right)$, where $f$ is an activation function.  The state of the reservoir $\mathbf{x}^{(p)}$ is then connected to an \textit{output layer} via a third weight matrix $\hat{\mathbf{y}}^{(p)}=\mathbf{c}\left(\mathbf{W}^{out}\mathbf{x}^{(p)}\right)$, where $c$ is the output function.  In this work, $c$ is an identity function, such that the output is given directly by the product of the output weight matrix and the reservoir state.  The output layer is trained such that the reservoir performs a particular function (e.g. regression, classification) of the inputs as
\begin{equation}
\mathbf{W}^{out*}=\operatorname*{arg\,min}_{\mathbf{W}^{out}} \frac{1}{2}\sum\limits_{p=1}^{m_{train}}\left(\mathbf{y}^{(p)}-\hat{\mathbf{y}}^{(p)}\right)^{2}
\end{equation}
where $\mathbf{y}^{(p)}$ is the expected output at timestep $p$ and $m_{train}$ is the size of the training set.  In this work, this optimization problem is solved via regularized least squares:
\begin{equation}
\mathbf{W}^{out*}=\left(\mathbf{X}^{\mathsf{T}}\mathbf{X}+\lambda\mathbf{I}\right)^{-1}\mathbf{X}^{\mathsf{T}}\mathbf{Y},
\label{eqn:linreg}
\end{equation}
where $\lambda$ is the regularization parameter, $\mathbf{I}$ is the identity matrix, $\mathbf{X}$, and $\mathbf{Y}$ are the matrices composed of the reservoir states and the expected outputs corresponding the training set $\mathbf{U}_{train}$, respectively.  Note that the only parameters that are modified during training are the output weights $\mathbf{W}^{out}$.  The random input and reservoir layers serve to randomly project the input into a high-dimensional space which increases the linear separability of inputs.  In addition, the recurrent connections of the reservoir provide a short-term memory that allows inputs to be integrated over time.  This is critical to analyzing data based on its behavior over multiple timesteps. 

\begin{figure}
\includegraphics[width=\columnwidth]{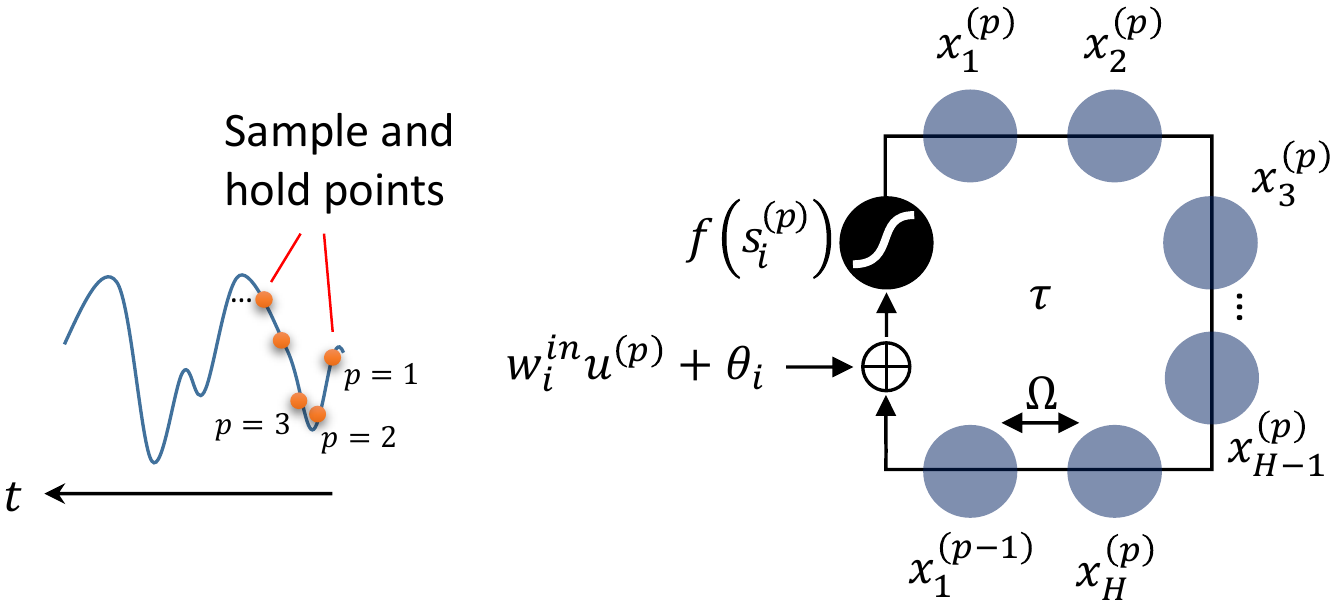}
\vspace{-5mm}
\caption{Overview of a TDR, showing the use of a delay line to create multiple virtual reservoir nodes from one physical node.}
\label{fig:tdr}
\end{figure}

A TDR (Figure \ref{fig:tdr}) is a special type of RC that shares resources in time to reduce routing overhead.  A single non-linear node (shown as an opaque black circle) provides the activation function, analogous to the sigmoid and spiking functions in the ESN and LSM, respectively.  The activation function can be any polynomial or transcendental function and is sometimes governed by a delay differential equation.  At each timestep $p$, the reservoir's input is sampled and held for a duration of $H$ smaller timesteps of duration $\Omega$.  For each of the $H$ timesteps, the held input is multiplied by an input weight $w_{i}^{in}$ and then added to a bias term $\theta_{i}$.  The weighted and biased input is added to the delayed state of the reservoir node, $x_{\tau-H}^{(p-1)}$, where $\tau\ge H$ is the delay of the feedback.  In this work, $\tau=H+1$.  The sum is then fed back into the activation function.  In this way, $H$ components make up the reservoir's state corresponding to each sampled and held input.  This approach is attractive because the hardware implementation usually consists of a simple circuit and a delay line without the routing overhead associated with ESNs and LSMs.  However, TDRs are more restricted in terms of their connectivity patterns and may require numerical simulation of DDEs when implemented in software.  For an instantaneous activation function (e.g. $f$ settles within $\Omega$) and $\tau=H+1$, a TDR has a unidirectional ring topology. 

\subsection{Stochastic Logic}
\label{section_stochastic_logic}

The next section will present an efficient hardware implementation of a TDR using stochastic computing techniques.  Stochastic logic was pioneered by von Neumann \cite{Neumann1956} half a century ago and was later adopted by the machine learning community to reduce hardware complexity, power, and unreliability \cite{Gaines1969,Brown2001}.  At the heart of stochastic computing is the stochastic representation of binary numbers.   In a single-line bipolar \cite{Brown2001} stochastic representation, a $q$-bit 2's complement number $z\in\{-2^{q-1},\ldots,-1,0,1,\ldots,2^{q-1}-1\}$ is mapped to a Bernoulli process $Z$ of
length $L$ \cite{Toral2000,Qian2011}:
\begin{equation}
\overline{z}\equiv\mathrm{Pr}(Z_{r}=1)=\frac{z+2^{q-1}}{2^{q}-1}=\frac{\tilde{z}+1}{2},
\label{eqn_px1}
\end{equation}
where the terms $\overline{z}\in[0,1]$ and $\tilde{z}\in[-1,1]$ are defined for convenience and $r=1,2,\ldots,L$ is an index into the stochastic bit stream.  Converting from the binary to the stochastic representation can be achieved using a random number generator such as a linear feedback shift register (LFSR) and a comparator.  If the random number is less than or equal to the value held in the register, then a logic 1 value is produced on the output.  Otherwise, the comparator output is logic 0 \cite{Toral2000}.  As $L$ becomes large, $\mathrm{Pr}(Z_{r}=1)$ approaches the value in (\ref{eqn_px1}).  Converting from a stochastic representation back to a digital number can be achieved by counting the number of 1's and 0's in the stream.  By initializing the counter to zero, adding `1' every time a `1' is encountered in the stream and subtracting `1' every time a `0' is encountered, the final counter value will be the 2's complement binary representation of the bit stream.

\begin{figure}[!t]
\centering
\includegraphics[width=\columnwidth]{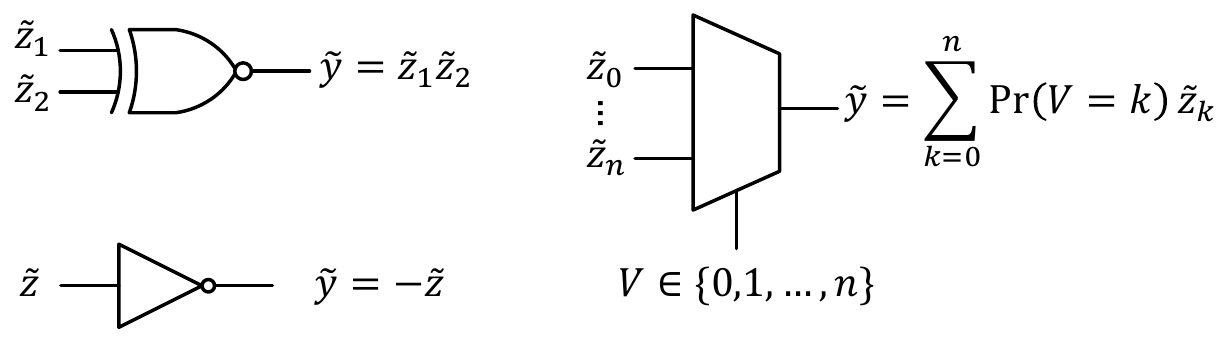}
\caption{Basic logic gates implementing mathematical operations on single-line bipolar stochastic bit steams.}
\label{fig:basicgates}
\end{figure}

One advantage of a stochastic representation is that several mathematical operations become trivial to implement in hardware.  For example, consider the logic function $Y_{i}=g(I_{1},I_{2},\ldots,I_{n})$, where each input $I_{j}$ to the function is mapped to a stochastic bit stream $Z^{j}$ and, therefore, the output is also a stochastic bit stream.  The probability that the function evaluates to `1' is given by
\begin{equation}
\mathrm{Pr}(Y_{i}=1)= \sum\limits_{I_{1},I_{2}\ldots} f(I_{1},I_{2},\ldots)\prod\limits_{k=1}^{n}\mathrm{Pr}(Z^{k}_{i}=I_{k}),
\label{eqn_poly_stoch}
\end{equation}
which is a multivariate polynomial in $\overline{z}_{1},\overline{z}_{2},\ldots,\overline{z}_{n}$ with integer coefficients and powers no greater than 1.  Implementations for a number of stochastic logic operations are shown in Figure \ref{fig:basicgates}.  Note that, in general, the implementation of stochastic logic operations will be different for unipolar representations.  For example, in the case of a bipolar representation, multiplication is implemented using an XNOR gate (see Figure \ref{fig:basicgates}).  However, for a unipolar representation, the same operation uses an AND gate.  Other basic operations such as negation and weighted averaging are achieved using inverters and multiplexers, respectively.

In addition to the simple mathematical operations shown in Figure \ref{fig:basicgates}, it will also be necessary for the stochastic logic RC to implement a non-linear activation function.  As indicated by (\ref{eqn_poly_stoch}), this is trivial if the activation function is a polynomial.  However, if the activation function is not a transcendental function, then one way to implement it is to approximate it with a polynomial.  Bernstein polynomials are a good choice, since they can approximate any function on the unit interval (or any other interval) with arbitrary precision, which was shown by Bernstein as part of a proof the Weierstrass approximation theorem \cite{Levasseur1984,Lorentz1986}.  The Bernstein basis polynomials of degree $n$ are defined as \cite{Lorentz1986}
\begin{equation}
b_{k,n}(\overline{z})\equiv {n\choose k}\overline{z}^{k}(1-\overline{z})^{n-k},\hspace{2mm}k=0,1,\ldots,n.
\end{equation}
A Bernstein polynomial of degree $n$ is defined as a linear combination of the $n^{\mathrm{th}}$-degree Bernstein basis polynomials:
\begin{equation}
B_{n}(\overline{z})\equiv\sum\limits_{k=0}^{n}\beta_{k}b_{k,n}(\overline{z})
\label{eqn_bernstein}
\end{equation}
The coefficients $\beta_{k}$ are called the Bernstein coefficients.  Furthermore, the $n^{\mathrm{th}}$-degree Bernstein polynomial for a function $f(\overline{z})$ is defined as
\begin{equation}
B_{n}(f;\overline{z})=\sum\limits_{k=0}^{n}f\left(\frac{k}{n}\right)b_{k,n}(\overline{z}).
\label{eqn_bernstein_approx_orig}
\end{equation}
Bernstein showed that $B_{n}(f;\overline{z})$ approaches $f(\overline{z})$ uniformly on $[0,1]$ as $n$ approaches infinity.

It can also be shown that the set of Bernstein basis polynomials $\{b_{k,n}\}$ of degree $n$ forms a basis for the space of power-form polynomials with real coefficients and degree no more than $n$.  In other words, the power-form polynomial
\begin{equation}
p(\overline{z})=\sum\limits_{i=0}^{n}a_{i}\overline{z}^{i}
\label{eqn_power_poly}
\end{equation}
can be written in the form of (\ref{eqn_bernstein}).  In \cite{Qian20112}, it is shown that the Bernstein coefficients can be obtained from the power-form coefficients as
\begin{equation}
\beta_{k}=\sum\limits_{i=0}^{k}{k\choose j}{n\choose j}^{-1}a_{i}.
\label{eqn_bernstein_poly}
\end{equation}
It is important to note that if $f$ in (\ref{eqn_bernstein}) maps the unit interval to the unit interval, then $f(k/n)$ is also in the unit interval.  Similarly, Qian et al. have proven that if $p$ in (\ref{eqn_power_poly}) maps the unit interval to the unit interval, then the Bernstein coefficients in (\ref{eqn_bernstein_poly}) will also be in the unit interval.  Coefficients in the unit interval are important because they can be represented stochastically.  

In summary, any non-polynomial (polynomial) function that maps the unit interval to the unit interval can be approximated by (written in the form of) a Bernstein polynomial with coefficients that are also in the unit interval.  To find the coefficients for non-polynomial functions, one may form the constrained optimization problem \cite{Qian2011}:
\begin{equation}
\begin{aligned}
& \underset{\{\beta_{0}\ldots\beta_{n}\}}{\text{minimize}}
& &
\int\limits_{0}^{1}\left(f(\overline{z})-\sum\limits_{k=0}^{n}\beta_{k}b_{k,n}\left(\overline{z}\right)\right)^{2}\mathrm{d}\overline{z}\\
& \text{subject to} & & \beta_{k}\in[0,1]\forall k=0,1,\ldots,n 
\end{aligned}
\end{equation}
which can be solved using numerical techniques.

Bernstein polynomials can be implemented in stochastic logic using only an adder and a MUX \cite{Qian2011}.  Consider an adder with $n$ inputs, each one a stochastic bit stream $Z_{1}\ldots S_{n}$.  Furthermore, let each bit stream be independent and identically distributed, such that $\overline{z}\equiv\overline{z}_{i}=\overline{z}_{j}\forall i,j$.  Then, at a particular time $t$, the adder will be adding $n$ bits, each one being `1' with probability $\overline{z}$.  Therefore, for their sum $V$ to be a particular value $k$ requires that $k$ bits are `1' and $n-k$ bits are `0'.  The probability of this occurring is ${n\choose k}\overline{z}^{k}(1-\overline{z})^{n-k}$.  Now, connecting the adder's output to the select line of a MUX, with inputs equal to the Bernstein coefficients, results in the Bernstein polynomial in (\ref{eqn_bernstein}).  

\section{Stochastic Logic TDR Design}
\label{section:design}

\begin{figure*}
\centering
\includegraphics[width=\textwidth]{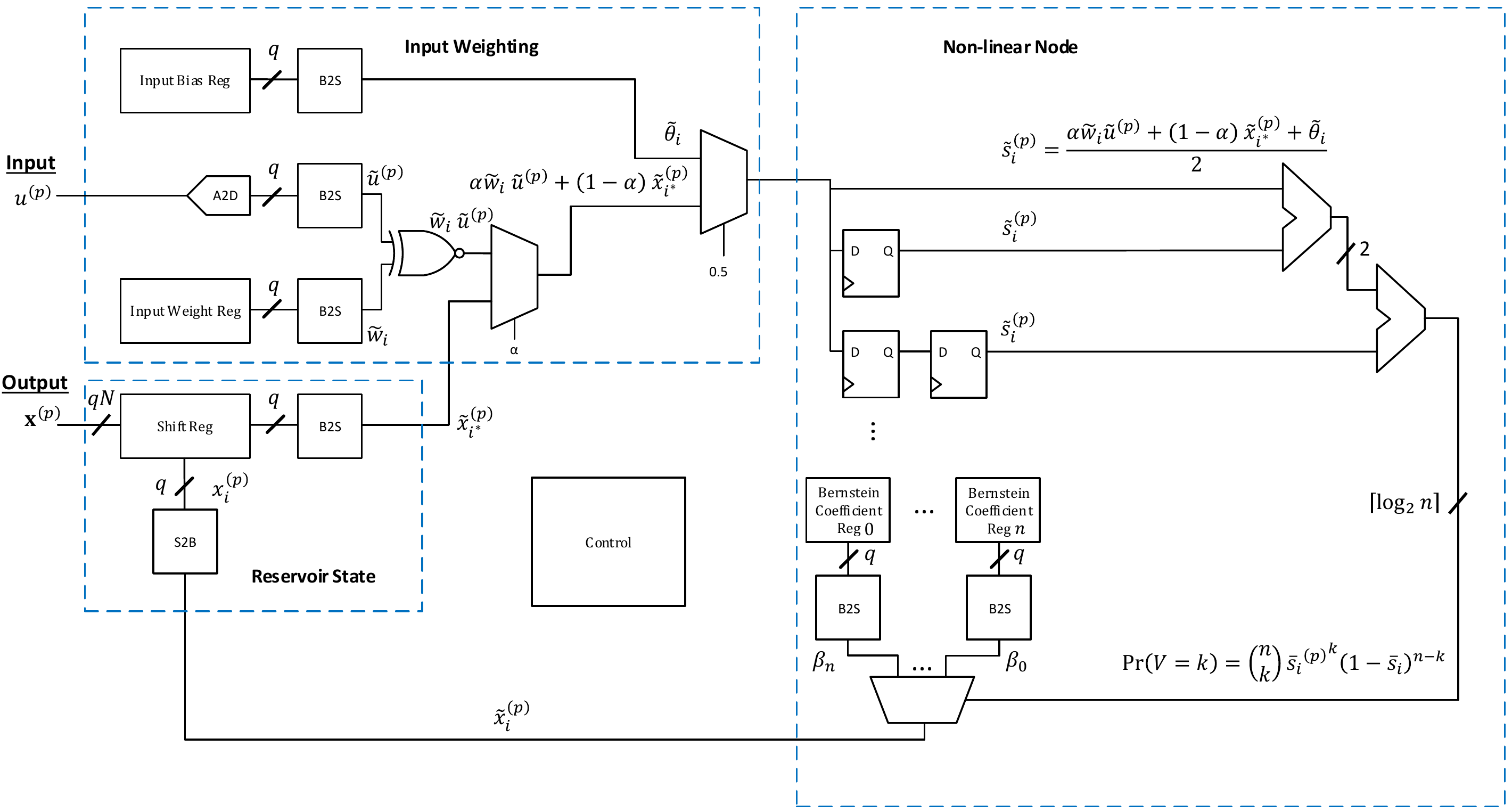}
\caption{Stochastic logic TDR designed in this work.}
\label{fig:digitdr}
\end{figure*}

\begin{figure}
\centering
\includegraphics[width=0.75\columnwidth]{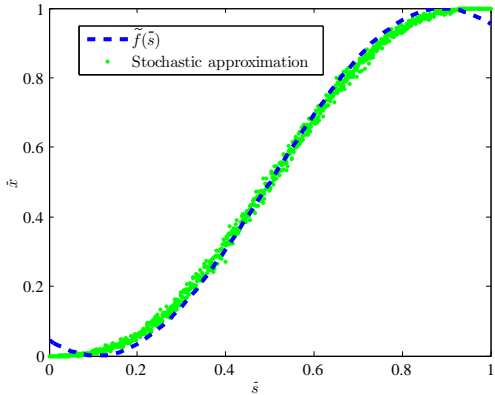}
\caption{Activation function and stochastic approximation implemented using Bernstein polynomials ($n=5$, $L=1000$).}
\label{fig:nlnodesim}
\end{figure}

The stochastic logic TDR designed in this work is shown in Figure \ref{fig:digitdr}.  The design is composed of three parts to provide input weight, compute the non-linear activation function, and hold the reservoir state.  The input weighting stage takes in an analog signal, converts it to a digital signal using an analog-to-digital (A2D) converter, and converts that to a stochastic bit stream using a binary-to-stochastic (B2S) converter.  In this design, the number of bits in the LFSRs in each B2S is equal to the number of bits in the binary representation of the input, $q$.  The stochastic representation of the input is then multiplied by the input weight using an XNOR gate, as discussed in the last section.  Then, the signal is mixed with the delayed reservoir state and added to the input bias using MUXes.  The non-linear node estimates the non-linear activation function $f\left(s\right)$ using Bernstein polynomials.  Shift registers are used to delay the non-linear node's input in order to create multiple statistically independent copies of the same stochastic bit stream.  In this design, the activation function implemented is
\begin{equation}
f(\tilde{s}) = \mathrm{sin}(\gamma \tilde{s}),
\end{equation}
where $\gamma$ is a frequency term.  However, recall from the discussion in Section \ref{section_stochastic_logic} that Bernstein polynomials map the unit interval to the unit interval.  By definition, $\overline{s}$ ranges from -1 to +1, and the $\mathrm{sin}$ function also ranges from -1 to +1.  In general, any function to be implemented by a Bernstein polynomial has to be shifted and scaled so that the portion of it that is used lies entirely in the unit square.  In the case of $\mathrm{sin}\left(\gamma \overline{s}\right)$, this is achieved by transforming the function as
\begin{equation}
\overline{f}\left(\overline{s}\right)=\frac{f\left(\Delta s\left[\overline{s}-0.5\right]\right)-f_{min}}{f_{max}-f_{min}},
\end{equation}
where $\Delta s$ is the domain of interest, $f_{max}$ is the maximum of the function on $[-\frac{\Delta s}{2},\frac{\Delta s}{2}]$, and $f_{min}$ is the minimum of the function on $[-\frac{\Delta s}{2},\frac{\Delta s}{2}]$.  For the function $f(\overline{s})$, this results in the dotted curve in Figure \ref{fig:nlnodesim}.  Also shown is the stochastic logic approximation using Bernstein polynomials with $L=1000$ and $n=5$.  Notice that the features of the curve around $\overline{s}=0$ and $\overline{s}=1$ are not reproduced well by this approximation but could be by increasing $n$.  However, it was found that the approximation, which is similar to a logistic sigmoid function works well for the benchmarks explored in this work.

\begin{figure*}
\centering
\subfigure[]{
\includegraphics[width=0.68\columnwidth]{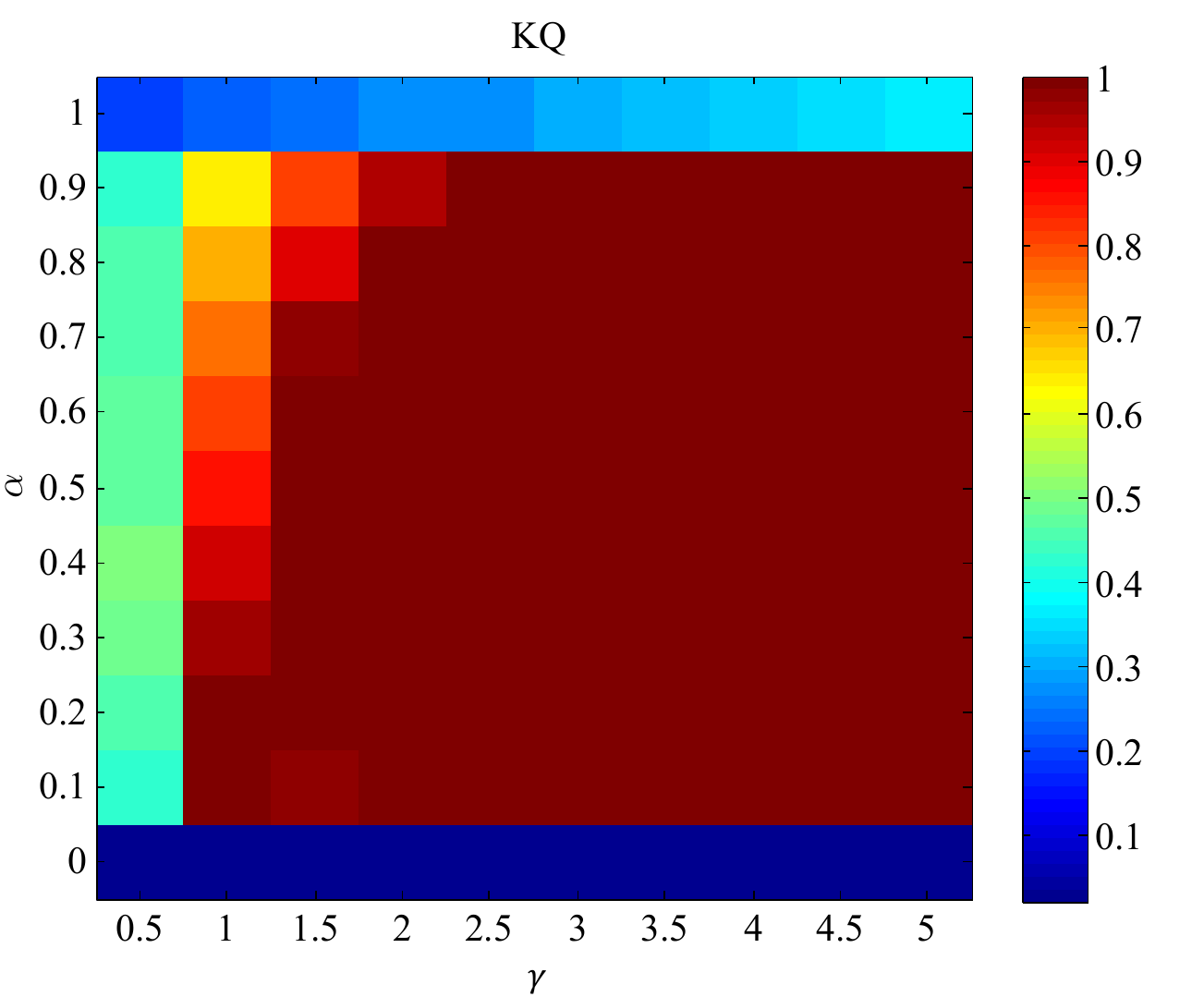}
}
\label{fig:l_infinity_kq}
\hspace{-5mm}
\subfigure[]{
\includegraphics[width=0.68\columnwidth]{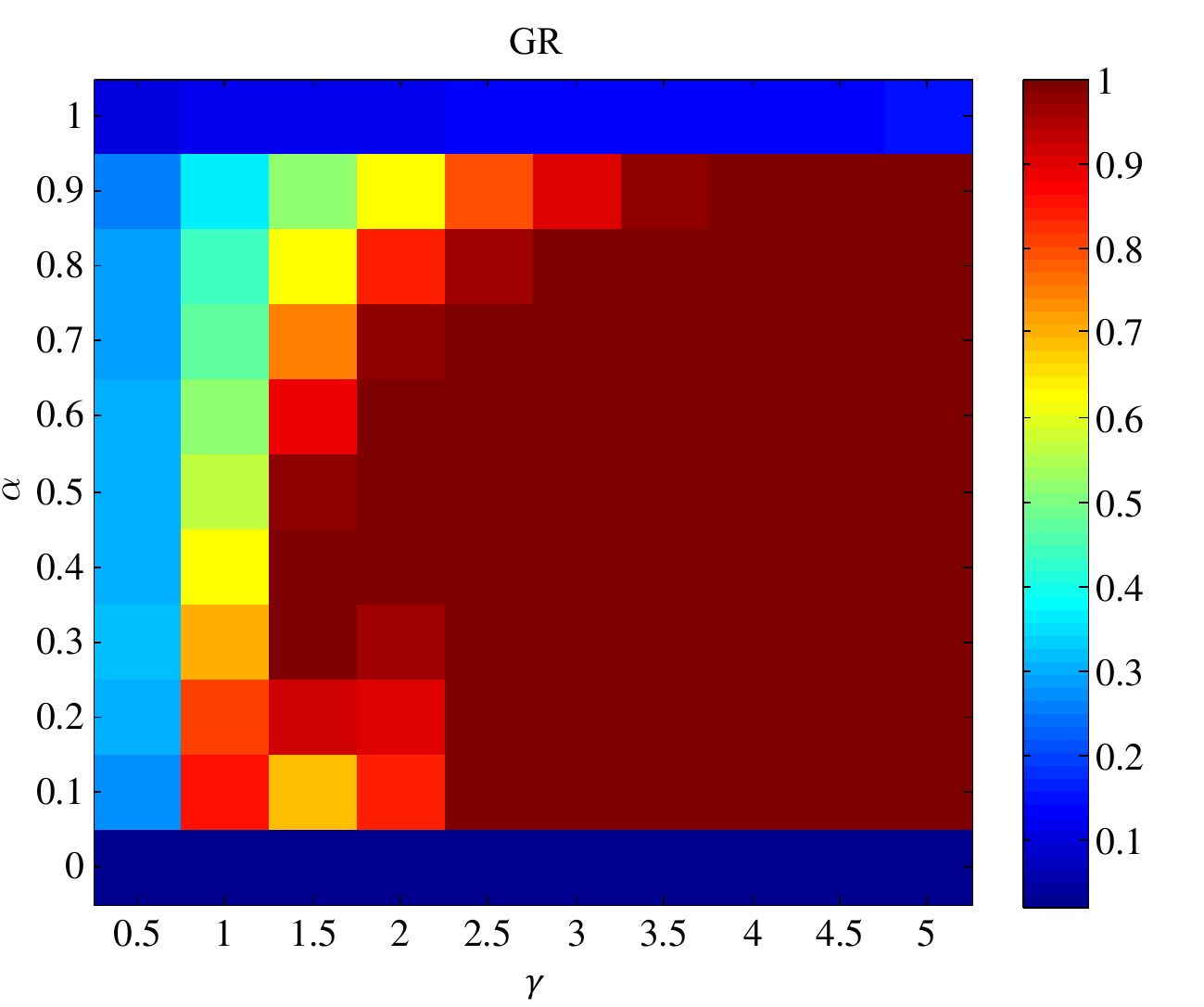}
\label{fig:l_infinity_gr}
}
\hspace{-5mm}
\subfigure[]{
\includegraphics[width=0.68\columnwidth]{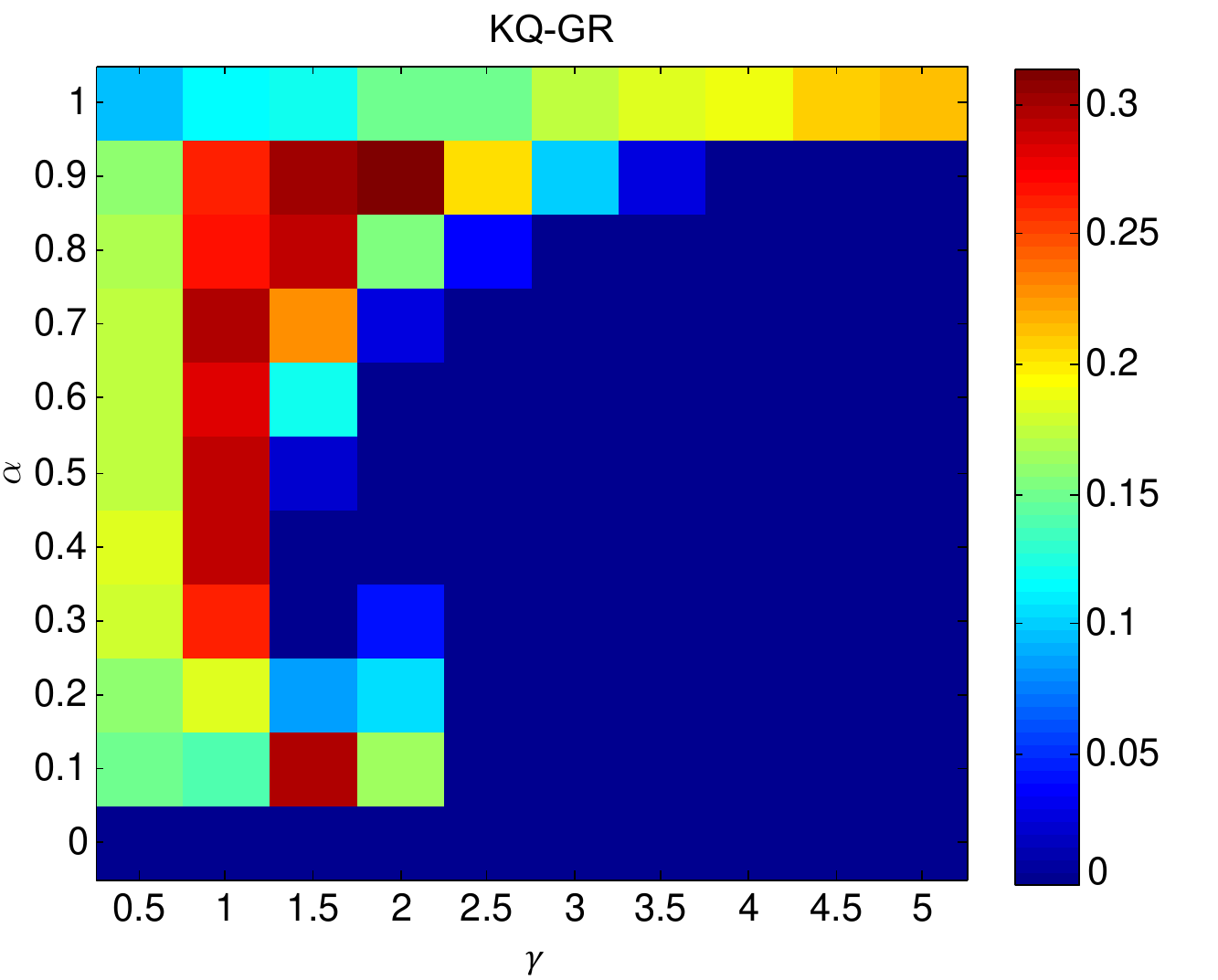}
\label{fig:l_infinity_kq_minus_gr}
}
\caption{Computational capability of the TDR for $L=\infty$. (a) KQ versus $\alpha$ and $\gamma$.  (b) GR vs. $\alpha$ and $\gamma$.  (c) KQ-GR vs. $\alpha$ and $\gamma$.  In all cases, $H=50$.}
\label{fig:metrics}
\end{figure*}

After the activation function is computed, the reservoir node $\overline{x}_{i}$ is converted back to a binary number using a stochastic-to-binary (S2B) converter and then placed in a shift register which holds the entire reservoir state $\mathbf{x}$.  Although the states could be stored in their stochastic representation, storing them as binary values is more area-efficient since $L\gg q$.  Note that a control block is also included in Figure \ref{fig:digitdr} to emphasize that the sample-and-hold circuit (inside the A2D), the B2S, and the S2B require a state machine when implemented in hardware.  However, all of the simulations in this work are behavioral and implemented in MATLAB, so this block wasn't explicitly required.  The shift register serves as the delay line shown in Figure \ref{fig:tdr}.  Note that training was not the focus of this work and performed using a non-stochastic implementation of (\ref{eqn:linreg}). 

Next is the task of choosing the design parameters $\alpha$ and $\gamma$.  One way is to look at application-dependent metrics such as accuracy, specificity, sensitivity, mean squared error, etc., and see how they vary over the parameter space via, e.g., a grid search.  Another way is to use metrics that capture features such as the reservoir's short-term memory capacity, ability to linearly separate input data, capability of mapping similar inputs to similar reservoir states (generalization), and different inputs to distant reservoir states (separation).  These types of application-independent metrics provide more insight into the effects of different parameter choices on the TDR's computing power than metrics like accuracy.  This work makes use of such metrics:  Kernel quality (KQ) and generalization rank (GR) \cite{legenstein2007edge}.  KQ is calculated by driving the reservoir with $H$ random input streams of length $m$.  At the end of each sequence, the reservoir's final state vector is inserted as a column into an $H\times H$ state matrix $\mathbf{X}$.  Then, KQ is equal to the rank of $\mathbf{X}$.  It is desired to have $\mathbf{X}$ be full rank, meaning that different inputs map to different reservoir states.  However, note that the number of training patterns is usually much larger than $H$, so if $\mathrm{rank}(\mathbf{X})=H$, it doesn't mean that any training dataset can be fit exactly.  In fact, exact fitting, or \textit{overfitting}, is generally bad, since it means that the TDR (or any machine learning algorithm) won't generalize well to novel inputs.  Therefore, another metric, GR, is used to measure the TDR's generalization capability.  GR is calculated in a similar way, except that all of the input vectors are identical except for some random noise.  GR is an estimate of the Vapnik-Chervonenkis dimension \cite{legenstein2007edge}, which is a measure of learning capacity.  A low GR is desirable.  Therefore, to achieve good performance, the difference KQ-GR should be maximized.  

Figure \ref{fig:metrics} shows the KQ and GR metrics for the stochastic logic TDR with $L=n=\infty$.  The values are normalized to $H$ and are averaged over 10 runs.  In each subplot, the size of the reservoir is $H=50$, and $m=50$.  KQ is close to 0 for $\alpha=0$.  When $\alpha=0$, the TDR does not accept any new inputs, so if the initial TDR state is all zeros, then the final state matrix will be a zero matrix, which has a rank of zero.  For larger $\alpha$ values, KQ becomes non-zero.  When $\gamma$ is small, the activation function is approximately linear, which leads to a smaller KQ.  In fact, it is likely that the TDR operates in the deterministic phase for $\gamma<1$.  As $\gamma$ becomes larger, KQ becomes equal to $H$.  This is because the non-linearity of the activation function increases with $\gamma$, and results in the TDR operating within the chaotic regime.  The GR metric (Figure \ref{fig:l_infinity_gr}) has similar behavior. When the difference KQ-GR is taken (Figure \ref{fig:l_infinity_kq_minus_gr}), a small region of optimal $\alpha$ and $\gamma$ is observed.  Values of $\alpha$ should be somewhere between 0 and 1.  If $\alpha$ is too large, then the TDR will have no memory, and if $\alpha$ is too small, then it will ignore inputs.  Furthermore, if $\gamma$ is too large, then the TDR will overfit the training data, and if $\gamma$ is too small then the TDR won't have enough dynamic behavior.  Also note that the optimal parameter values will have some application dependence.  In this work, parameter values of $\alpha=0.2$ and $\gamma=2$ were determined empirically to be the best for the studied benchmarks.  However, from Figure \ref{fig:l_infinity_kq_minus_gr}, it appears that this choice is suboptimal.  Therefore, one should be cautious when using metrics such as GR and KQ and always consider the behavior of the RC for a chosen set of applications.

\begin{figure}
\centering
\includegraphics[width=0.75\columnwidth]{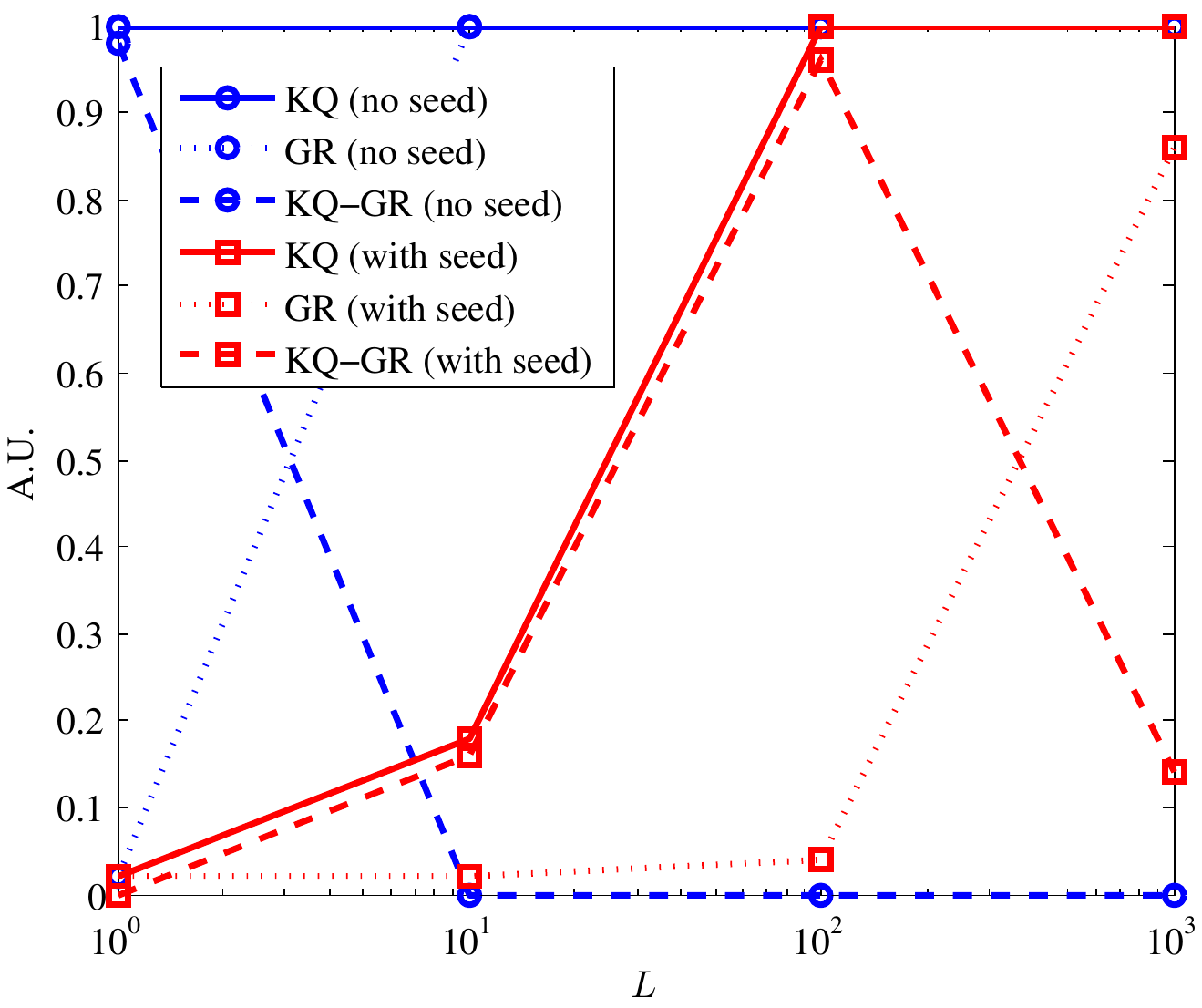}
\caption{KQ and GR vs. $L$ in the stochastic logic TDR.  Results are shown for the cases where the PRNG isn't (no seed) and is (seed) re-seeded for each reservoir node.}
\label{fig:kqandgrvsl}
\end{figure}

Studied next was the effect of the stochastic bit stream length $L$ on the TDR metrics.  Intuitively, one would expect that a small value of $L$ would lead to both a large KQ and GR, since the variance introduced from the stochastic computation is $\propto 1/L$.  Indeed, this is true.  Figure \ref{fig:kqandgrvsl} shows the KQ and GR metrics for two cases.  In the first case (no seed), each LFSR was only seeded at the beginning of the simulation.  This resulted in KQ-GR equal to zero over all $L$ values, except $L=1$, where the noise wasn't large enough to modify the stochastic representation.  From the previous discussion, we see that KQ-GR will eventually become non-zero as $L\rightarrow\infty$.  However, that would mean that the TDR may have to wait an impractical number of clock cycles for each calculation.  Instead, the approach used in this work is to re-seed each PRNG for every reservoir node, with a unique seed for that node.  Although this doesn't eliminate stochastic noise, it does keep the effect of the noise approximately constant for each node.  With re-seeding (Figure \ref{fig:kqandgrvsl}), KQ-GR is non-zero for reasonable $L$ values such as 100 and 1000.

\section{Benchmark Results and Analysis}
\label{section:results}

\begin{figure}
\centering
\includegraphics[width=0.75\columnwidth]{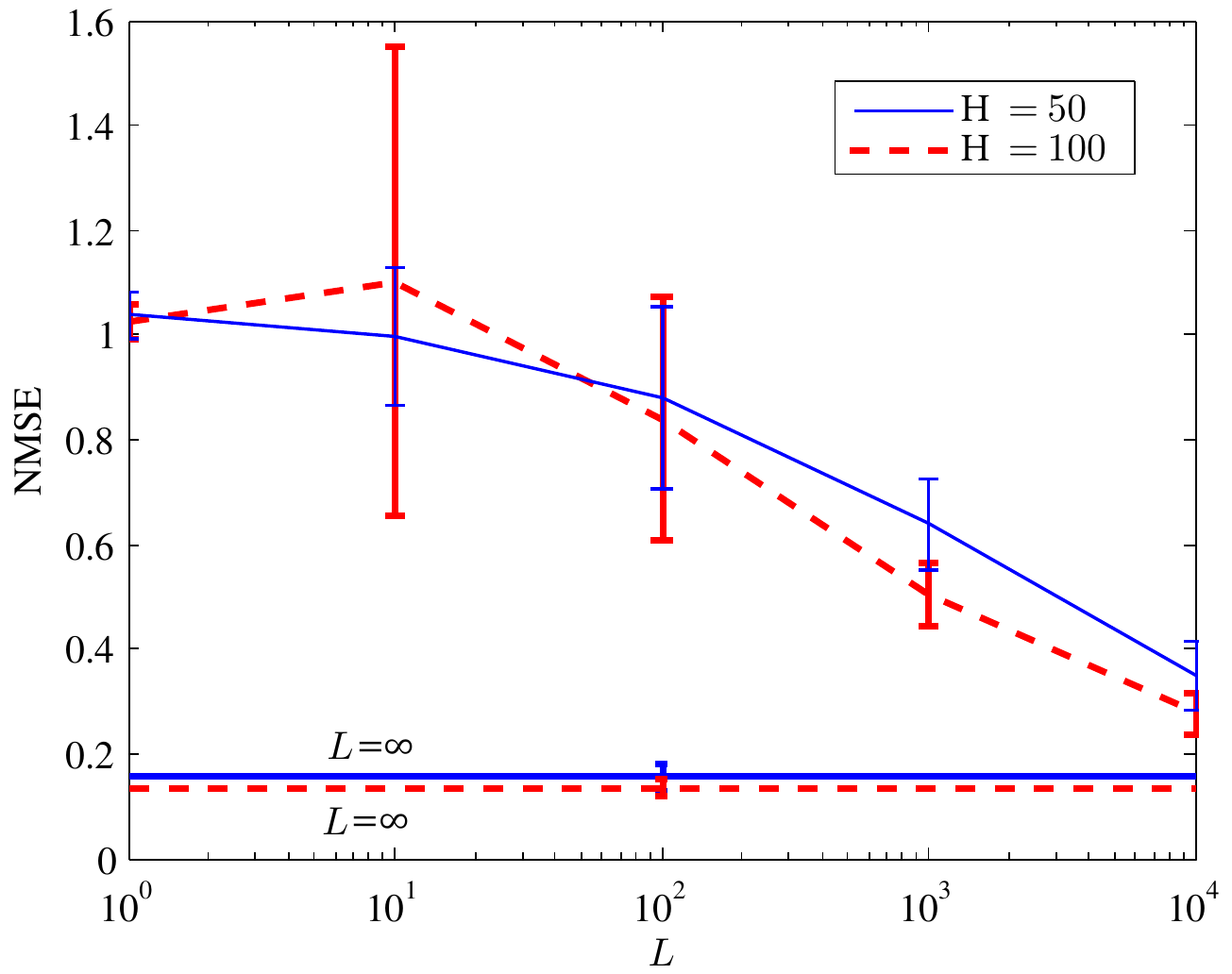}
\caption{NARMA10 benchmark NMSE vs. $L$ using stochastic logic TDR for reservoir sizes of 10 and 100.  Lines marked $L=\infty$ correspond to ideal implementations with no stochastic noise.}
\label{fig:narma10vsL}
\end{figure}

The stochastic logic TDR proposed in this work was tested on two simple benchmarks: NARMA10 (regression) and sine/square wave discrimination (classification).  NARMA10 is a standard benchmark used in RC research.  Given a random vector $\mathbf{u}\in [0,0.5]^{m}$, the goal is to train the RC to compute
\begin{equation}
y^{(p+1)}=0.3y_{i}^{(p)}+0.05y^{(p)}\sum\limits_{k=0}^{9}y^{(p-k)}+1.5u^{(p)}u^{(p-9)}+0.1.
\end{equation}
In this work, the TDR was trained on a set of 1000 datapoints and tested on an additional 1000.
Figure \ref{fig:narma10vsL} shows the normalized mean square error (NMSE) of the TDR on the test data.  The NMSE is calculated as
\begin{equation}
\mathrm{NMSE}=\frac{\sum_{p}\left(y^{(p)}-\hat{y}^{(p)}\right)^{2}}{\sum_{p}\left(y^{(p)}-\langle \mathbf{y}_{train} \rangle\right)},
\end{equation}
where $\langle\cdot\rangle$ is the arithmetic mean, and $\mathbf{y}_{train}$ is the vector of the entire training sequence.  The plot shows an average over 10 runs, with error bars indicating the standard deviation.  It is observed that the NMSE is much larger than the ``ideal" case ($L=\infty$) until $L$ becomes very large (e.g. $1\times 10^{4}$).  Such a large value of $L$ would give the TDR a prohibitively large latency and may only be feasible if the time constant of the input signal is very large (i.e. the input changes slowly).  It is possible to improve the NMSE by adding more reservoir nodes, which can be observed as $H$ changes from 50 to 100.  However, in a digital implementation, each new node requires additional hardware to store the additional component of the reservoir state.  This could become costly if $H$ is very large.

\begin{figure}
\centering
\includegraphics[width=0.75\columnwidth]{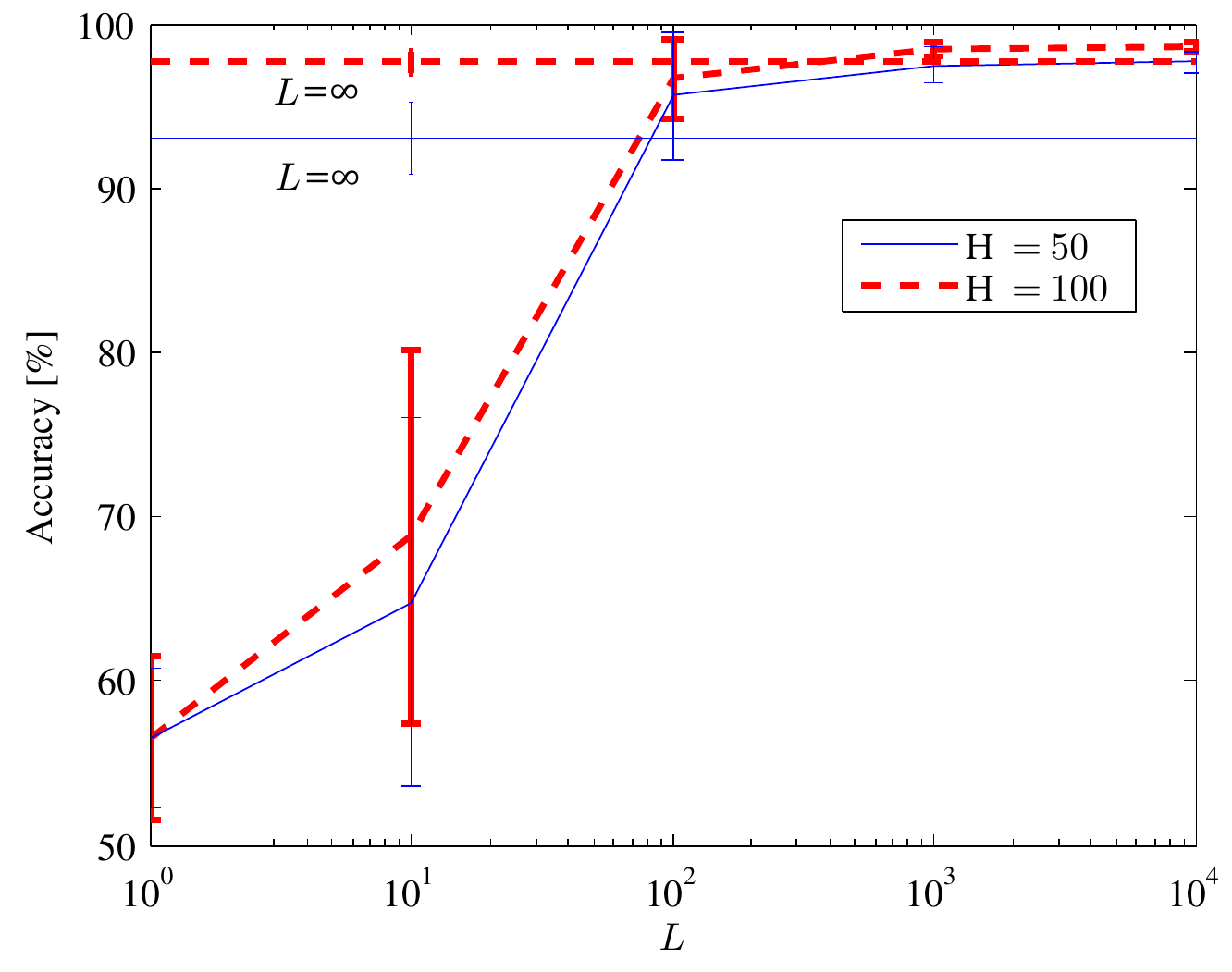}
\caption{Sine/square wave discrimination benchmark accuracy vs. $L$ using stochastic logic TDR for reservoir sizes of 10 and 100.  Lines marked $L=\infty$ correspond to ideal implementations with no stochastic noise.}
\label{fig:sinesquare}
\end{figure}

To test its performance on classification tasks, the stochastic logic TDR was trained to discriminated between sine and square wave signals.  1000 training and test cases were used, as was the case for the NARMA 10 benchmark.  The training and test sequences were created by randomly interposing segments of sine waves into a square wave such that there was a 50\% chance that any point in the sequence was part of a sine (or square) wave.  The results are shown in Figure \ref{fig:sinesquare}.  As expected, at very small values of $L$, the TDR gives classification accuracies that are close to random chance.  However, when $L$ is equal to 100, which is fairly small, the TDR performs approximately as well as the deterministic TDR.  In-fact, for $H=50$, the accuracy of the stochastic TDR surpasses that of the deterministic design.  At first, this seems counterintuitive, since smaller values of $L$ result in more noise.  However, the noise is actually acting as a regularizer in this case, reducing the stochastic TDR's ability to overfit.

\section{Conclusions and Future Work}
\label{section:conclusions}

This work studied a novel TDR design that uses stochastic logic to perform weighting, biasing, and activation function operations.  The design is more flexible than previous approaches as it allows any activation function to be implemented after it is properly shifted and scaled.  Optimal design parameters are chosen based on kernel quality and generalization rank, and a method for reducing stochastic noise using re-seeding was proposed.  The design was tested using the NARMA10 and sine/square wave discrimination benchmarks.  Results indicate that high-precision benchmarks, such as NARMA10, do not perform well on a stochastic TDR due to random noise.  However, for classification benchmarks, the stochastic TDR is more area efficient than previous design approaches.  This paper provides a foundation for future research directions related to stochastic logic TDRs.  Some potentially fruitful avenues include investigation of other activation functions (e.g. Mackey-Glass), methods for reducing the complexity of B2S converters (i.e. removing expensive comparators and LFSRs), exploring methods for reducing stochastic noise, and investigation of emerging memory technologies for more efficient storage of reservoir states.

\section*{Acknowledgments}

The author is grateful to Nathan McDonald, Clare Thiem, Lisa Loomis, and Ashley Prater for proofreading the manuscript and providing helpful discussions.

The material and results presented in this paper have been CLEARED (Distribution A) for public release, unlimited distribution by AFRL, case number 88ABW-2016-6393. Any opinions, findings and conclusions or recommendations expressed in this material are those of the author and do not necessarily reflect the views of AFRL or its contractors.

\bibliographystyle{IEEEtran}
\bibliography{test}

\begin{thebibliography}{10}
\providecommand{\url}[1]{#1}
\csname url@samestyle\endcsname
\providecommand{\newblock}{\relax}
\providecommand{\bibinfo}[2]{#2}
\providecommand{\BIBentrySTDinterwordspacing}{\spaceskip=0pt\relax}
\providecommand{\BIBentryALTinterwordstretchfactor}{4}
\providecommand{\BIBentryALTinterwordspacing}{\spaceskip=\fontdimen2\font plus
\BIBentryALTinterwordstretchfactor\fontdimen3\font minus
  \fontdimen4\font\relax}
\providecommand{\BIBforeignlanguage}[2]{{%
\expandafter\ifx\csname l@#1\endcsname\relax
\typeout{** WARNING: IEEEtran.bst: No hyphenation pattern has been}%
\typeout{** loaded for the language `#1'. Using the pattern for}%
\typeout{** the default language instead.}%
\else
\language=\csname l@#1\endcsname
\fi
#2}}
\providecommand{\BIBdecl}{\relax}
\BIBdecl

\bibitem{jaeger2001echo}
H.~Jaeger, ``The “echo state” approach to analysing and training recurrent
  neural networks-with an erratum note,'' \emph{Bonn, Germany: German National
  Research Center for Information Technology GMD Technical Report}, vol. 148,
  p.~34, 2001.

\bibitem{maass2002real}
W.~Maass, T.~Natschl{\"a}ger, and H.~Markram, ``Real-time computing without
  stable states: A new framework for neural computation based on
  perturbations,'' \emph{Neural computation}, vol.~14, no.~11, pp. 2531--2560,
  2002.

\bibitem{Woodward2011}
A.~Woodward and T.~Ikegami, ``A reservoir computing approach to image
  classification using coupled echo state and back-propagation neural
  networks,'' in \emph{Proc. of 26th Int. Conf. on Image and Vision Computing,
  Auckland, New Zealand, November}, 2011, pp. 543--458.

\bibitem{Kudithipudi2015}
D.~Kudithipudi, Q.~Saleh, C.~Merkel, J.~Thesing, and B.~Wysocki, ``Design and
  analysis of a neuromemristive reservoir computing architecture for biosignal
  processing,'' \emph{Frontiers in neuroscience}, vol.~9, 2015.

\bibitem{Tsai2010}
\BIBentryALTinterwordspacing
C.-Y. Tsai, X.~Dutoit, K.-T. Song, H.~{Van Brussel}, and M.~Nuttin, ``{Robust
  face tracking control of a mobile robot using self-tuning Kalman filter and
  echo state network},'' \emph{Asian Journal of Control}, vol.~12, no.~4, pp.
  488--509, 2010. [Online]. Available:
  \url{http://doi.wiley.com/10.1002/asjc.204}
\BIBentrySTDinterwordspacing

\bibitem{Appeltant2011}
L.~Appeltant, M.~C. Soriano, G.~Van~der Sande, J.~Danckaert, S.~Massar,
  J.~Dambre, B.~Schrauwen, C.~R. Mirasso, and I.~Fischer, ``Information
  processing using a single dynamical node as complex system,'' \emph{Nature
  communications}, vol.~2, p. 468, 2011.

\bibitem{vinckier2015high}
Q.~Vinckier, F.~Duport, A.~Smerieri, K.~Vandoorne, P.~Bienstman, M.~Haelterman,
  and S.~Massar, ``High-performance photonic reservoir computer based on a
  coherently driven passive cavity,'' \emph{Optica}, vol.~2, no.~5, pp.
  438--446, 2015.

\bibitem{Haynes2015}
N.~D. Haynes, M.~C. Soriano, D.~P. Rosin, I.~Fischer, and D.~J. Gauthier,
  ``{Reservoir computing with a single time-delay autonomous Boolean node},''
  \emph{Physical Review E - Statistical, Nonlinear, and Soft Matter Physics},
  vol.~91, no.~2, pp. 1--5, 2015.

\bibitem{jin2016sso}
Y.~Jin, Y.~Liu, and P.~Li, ``{SSO-LSM}: A sparse and self-organizing
  architecture for liquid state machine based neural processors,'' in
  \emph{Nanoscale Architectures (NANOARCH), 2016 IEEE/ACM International
  Symposium on}.\hskip 1em plus 0.5em minus 0.4em\relax IEEE, 2016, pp. 55--60.

\bibitem{Alomar2015}
M.~L. Alomar, M.~C. Soriano, M.~Escalona-Mor{\'a}n, V.~Canals, I.~Fischer,
  C.~R. Mirasso, and J.~L. Rossell{\'o}, ``Digital implementation of a single
  dynamical node reservoir computer,'' \emph{IEEE Transactions on Circuits and
  Systems II: Express Briefs}, vol.~62, no.~10, pp. 977--981, 2015.

\bibitem{Alomar2016}
M.~L. Alomar, V.~Canals, N.~Perez-Mora, V.~Mart{\'{i}}nez-Moll, and J.~L.
  Rossell{\'{o}}, ``{FPGA-Based Stochastic Echo State Networks for Time-Series
  Forecasting},'' \emph{Computational Intelligence and Neuroscience}, vol.
  2016, no.~ND, pp. 1--14, 2016.

\bibitem{Verstraeten}
D.~Verstraeten, B.~Schrauwen, and D.~Stroobandt, ``Reservoir computing with
  stochastic bitstream neurons,'' in \emph{Proceedings of the 16th annual
  Prorisc workshop}, 2005, pp. 454--459.

\bibitem{Neumann1956}
J.~von Neumann, ``Probabilistic logics and the synthesis of reliable organisms
  from unreliable components,'' \emph{Automata Studies}, vol.~34, pp. 43--99,
  1956.

\bibitem{Gaines1969}
B.~Gaines, ``Stochastic computing systems,'' in \emph{Advances in Information
  Systems Science}, J.~Tou, Ed.\hskip 1em plus 0.5em minus 0.4em\relax Plenum,
  1969, ch.~2, pp. 37--172.

\bibitem{Brown2001}
B.~Brown and H.~Card, ``Stochastic neural computation. {I}. {C}omputational
  elements,'' \emph{IEEE Transactions on Computers}, vol.~50, no.~9, pp.
  891--905, Sep. 2001.

\bibitem{Toral2000}
S.~Toral, J.~Quero, and L.~Franquelo, ``Stochastic pulse coded arithmetic,'' in
  \emph{IEEE International Symposium on Circuits and Systems}, ser. ISCAS 2000,
  vol.~1, 2000, pp. 599--602.

\bibitem{Qian2011}
W.~Qian, X.~Li, M.~Riedel, K.~Bazargan, and D.~Lilja, ``An architecture for
  fault-tolerant computation with stochastic logic,'' \emph{IEEE Transactions
  on Computers}, vol.~60, no.~1, pp. 93--105, Jan. 2011.

\bibitem{Levasseur1984}
K.~M. Levasseur, ``Bernstein polynomials,'' \emph{The American Mathematical
  Monthly}, vol.~91, no.~4, pp. 249--250, 1984.

\bibitem{Lorentz1986}
G.~G. Lorentz, \emph{Bernstein Polynomials}, 2nd~ed.\hskip 1em plus 0.5em minus
  0.4em\relax Chelsea Publishing Company, 1986.

\bibitem{Qian20112}
W.~Qian, M.~D. Riedel, and I.~Rosenberg, ``Uniform approximation and bernstein
  polynomials with coefficients in the unit interval,'' \emph{European Journal
  of Combinatorics}, vol.~32, no.~3, pp. 448--463, 2011.

\bibitem{legenstein2007edge}
R.~Legenstein and W.~Maass, ``Edge of chaos and prediction of computational
  performance for neural circuit models,'' \emph{Neural Networks}, vol.~20,
  no.~3, pp. 323--334, 2007.

\end{thebibliography}

\end{document}